\documentclass[10pt,twocolumn,letterpaper]{article}

\usepackage[T1]{fontenc}
\usepackage{times}
\usepackage{mathptmx}
\usepackage[letterpaper,top=1in,bottom=1in,left=0.8125in,right=0.8125in,columnsep=0.3125in]{geometry}
\usepackage{amsmath}
\usepackage{amssymb}
\usepackage{graphicx}
\usepackage{booktabs}
\usepackage{multirow}
\usepackage{adjustbox}
\usepackage{caption}
\usepackage{xcolor}
\usepackage{enumitem}
\usepackage{microtype}
\usepackage[numbers,sort&compress]{natbib}
\usepackage{url}
\definecolor{linkblue}{rgb}{0.21,0.49,0.74}
\usepackage[breaklinks,colorlinks,allcolors=linkblue]{hyperref}
\hypersetup{pdftitle={Measuring the Reasoning-Action Coupling in Driving Vision-Language-Action Models through Counterfactuals},pdfauthor={Gabriel Manalu, Saad Wazir, Thao Do, Seongah Kim, Hyeontaek Hwang, Tolcha Yalew Kidane, Daeyoung Kim}}

\makeatletter
\def\@maketitle{%
  \newpage\null\vskip .375in
  \begin{center}
    {\Large\bfseries \@title \par}
    \vskip .35in
    {\large \lineskip .5em
      \begin{tabular}[t]{c}\@author\end{tabular}\par}
    \vskip .5em
  \end{center}
  \par\vskip .5em}
\renewenvironment{abstract}{\centerline{\large\bfseries Abstract}\vspace*{12pt}\itshape\noindent\ignorespaces}{\par}
\def\@seccntformat#1{\csname the#1\endcsname.\hspace{0.5em}}
\renewcommand\section{\@startsection{section}{1}{\z@}%
  {-2.0ex plus -0.5ex minus -.2ex}{1.5ex plus 0.3ex minus .2ex}{\large\bfseries\raggedright}}
\renewcommand\subsection{\@startsection{subsection}{2}{\z@}%
  {-1.8ex plus -0.5ex minus -.2ex}{0.8ex}{\fontsize{11}{13}\selectfont\bfseries\raggedright}}
\makeatother
\newcommand{\afpair}[2]{$#1$\,/\,$#2$}

\newcommand{\chan}{reasoning}
\newcommand{\Chan}{Reasoning}

\DeclareRobustCommand{\micon}[1]{\raisebox{-0.2em}{\includegraphics[height=1.05em]{figures/basic_car/#1}}\,}

\newcommand{\autovlaS}{\micon{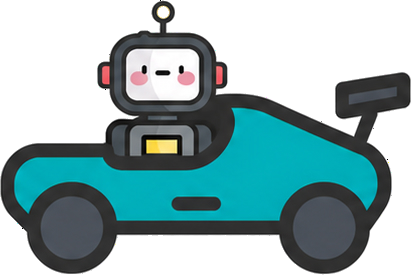}AutoVLA}

\newcommand{\alponeS}{\micon{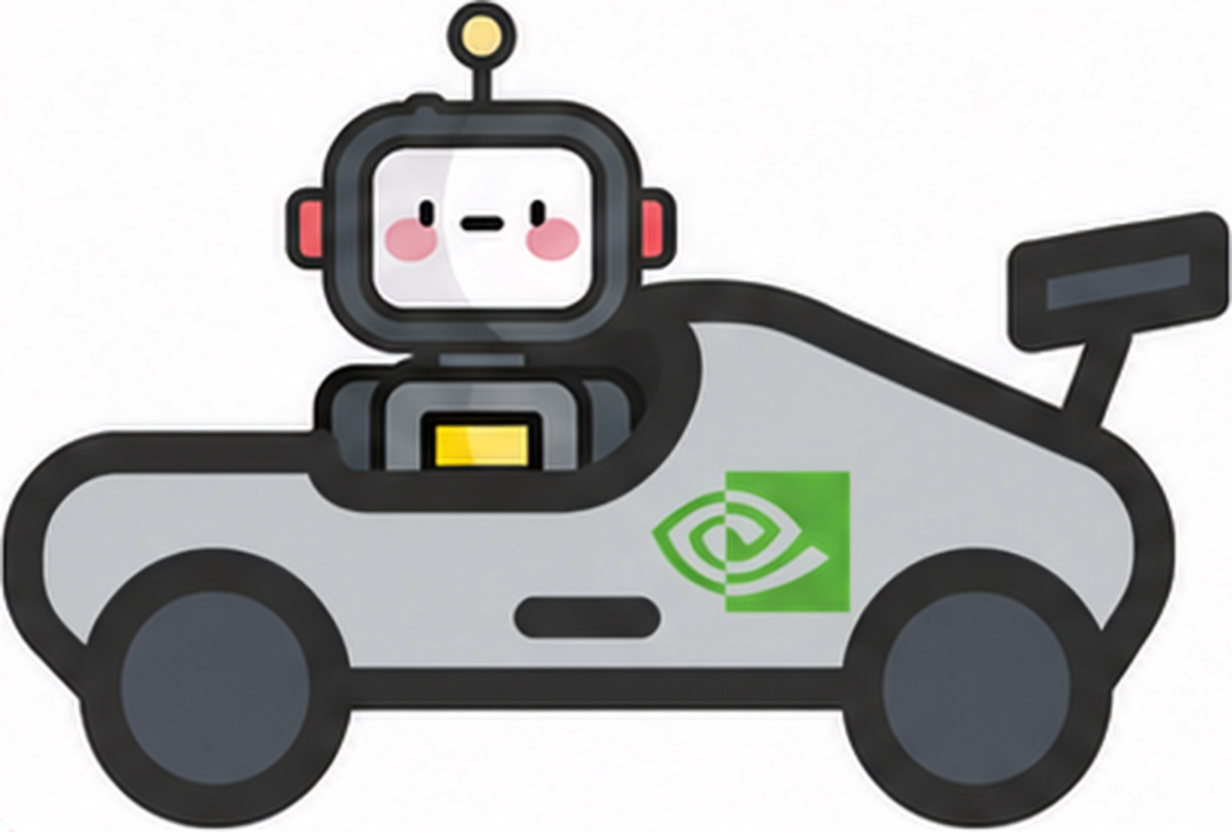}Alpamayo-1.5}

\newcommand{\alprS}{\micon{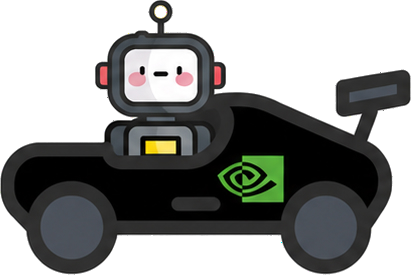}Alpamayo-R1}
\newcommand{\orionM}{\micon{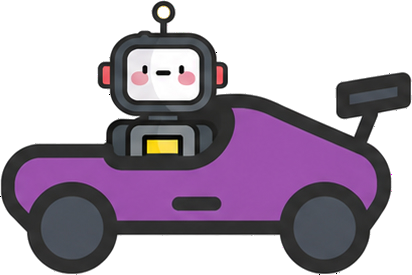}ORION}
\newcommand{\eraser}{\micon{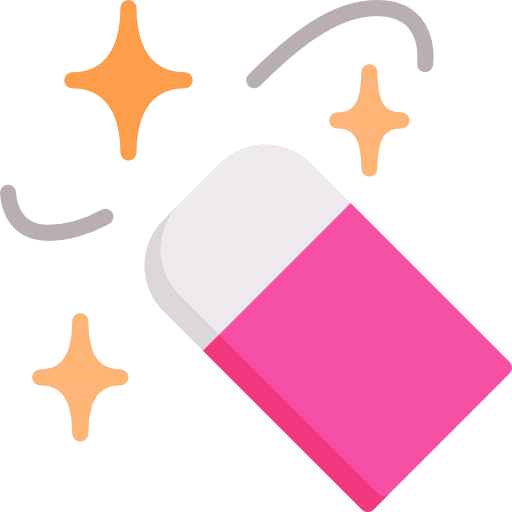}}

\newcommand{\mistakeop}{\textsc{mistake}}
\newcommand{\negateop}{\textsc{negate}}
\newcommand{\swapop}{\textsc{swap}}
\newcommand{\paraphrase}{\textsc{paraphrase}}
\newcommand{\filler}{\textsc{filler}}
\newcommand{\deleteop}{\textsc{delete}}
\newcommand{\expruni}{\noindent\textbf{Experiments}\ }
\newcommand{\resruni}{\noindent\textbf{Results}\ }

\newcommand{\fwsub}[1]{\subsection{#1}}
\newcommand{\dssub}[1]{\subsection{#1}}

\title{Measuring the Reasoning--Action Coupling in Driving\\Vision--Language--Action Models through Counterfactuals}

\author{%
Gabriel Manalu\quad
Saad Wazir\quad
Thao Do\quad
Seongah Kim\\
Hyeontaek Hwang\quad
Tolcha Yalew Kidane\quad
Daeyoung Kim\\[0.3em]
School of Computing, KAIST, Republic of Korea
}
\date{}
\begin{document}

\twocolumn[{%
\maketitle
\vspace{-0.4cm}

\noindent\begin{minipage}{\textwidth}
  \centering

  \includegraphics[width=\linewidth]{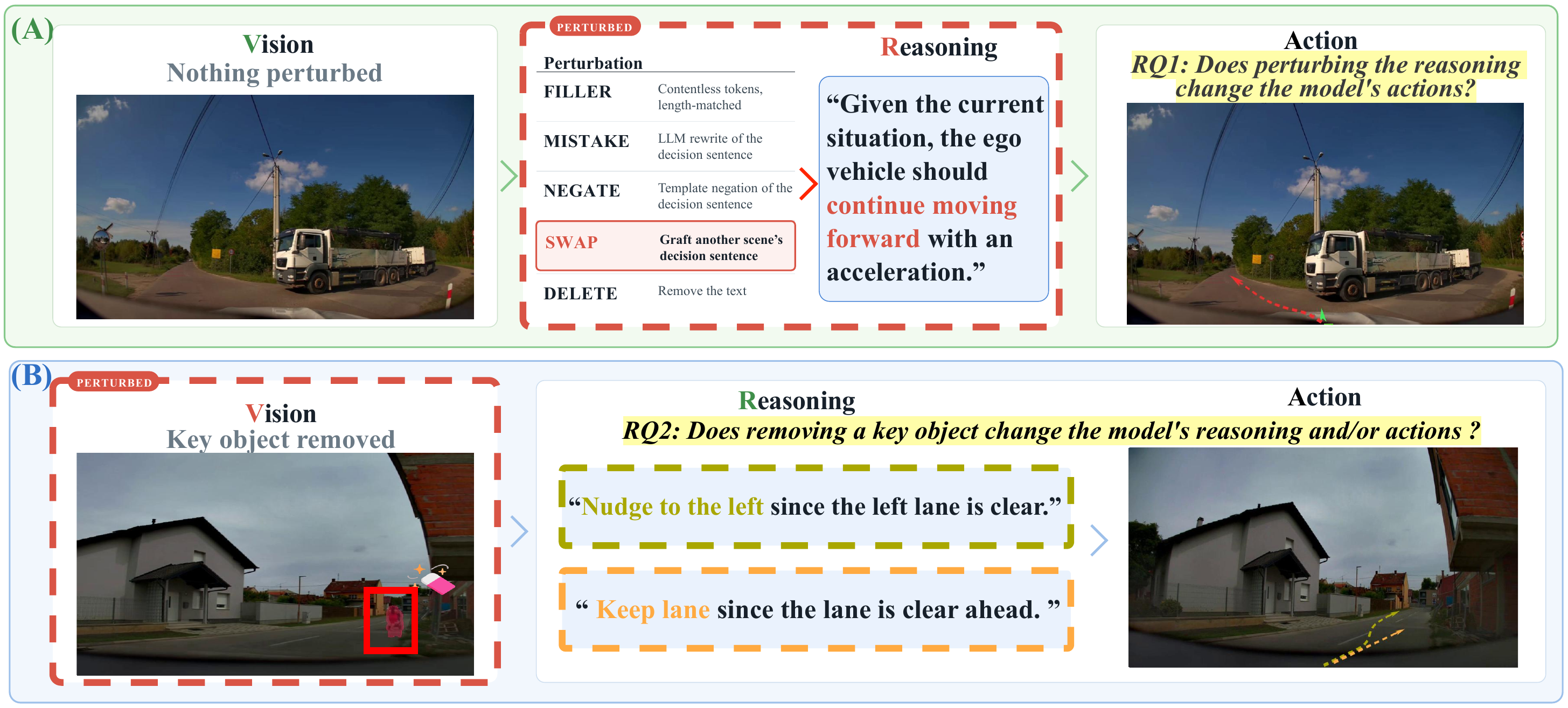}

  \vspace{0.2cm}

  \captionsetup{
    hypcap=false,
    font=small,
    labelfont=bf,
    justification=justified,
    singlelinecheck=false
  }
  \captionof{figure}{
    Driving VLAs present generated \chan{} as the account of why the car acts. Whether the trajectory depends on that \chan{} (\textbf{A}), and
    whether the \chan{} and actions are grounded on the image (\textbf{B}) have yet to be measured in the context of measuring the reasoning-action coupling of these VLAs.
  }
  \label{fig:motivation}
\end{minipage}

\vspace{0.15cm}
}]

\begin{abstract}
Driving vision--language--action (VLA) models bring an appealing form of explainability to a safety-critical domain, since a system that can state the reasons for its maneuvers offers the transparency that deployment and public trust in autonomous driving demand. These models read the world into language before they act on it, describing the scene and reasoning over it in text that ranges from chain-of-thought to multi-stage question answering on the way to a trajectory. That framing quietly assumes two separable properties, namely that the trajectory \emph{follows} the generated reasoning and that the generated reasoning \emph{tracks} the causes that actually produce the trajectory. To thoroughly investigate these properties, we apply causal, counterfactual interventions to four driving VLAs across different benchmarks. First, we perturb the models' own reasoning and show that altering its meaning produces small trajectory changes relative to blanking the same models' camera inputs, a pattern that persists across benchmarks and model architectures. We then construct scene-level counterfactuals through photorealistic removal of key objects named in the models' reasoning and analyze whether the generated reasoning and trajectories track these changes. Overall, our findings indicate that the reasoning these models present to account for their resulting driving behavior is weakly coupled to that behavior itself, pointing to limits on the explainability these systems are intended to provide.
\end{abstract}

\section{Introduction}
\label{sec:intro}

An autonomous driving model that writes ``a pedestrian is entering the crosswalk, so I will slow down'' before emitting a trajectory is making a double promise about its generated \chan{}. It claims that the stated reason contributes to the predicted maneuver, and that the pedestrian is a relevant cause of the vehicle slowing down. Most driving vision--language--action (VLA) models make some version of this promise. The text they emit ranges from multi-step chains of thought \citep{zhou2025autovla,yuan2025autodrive} and single-sentence chains of causation \citep{wang2025alpamayo} to question--answer exchanges about the scene \citep{tian2024drivevlm,sima2024drivelm}. Across these systems, generated \chan{} is presented as a means of explaining driving decisions \citep{xu2023drivegpt4,hwang2024emma,jiang2024senna,yuan2025autodrive,chi2025impromptu,hu2025vla}.

Existing evaluations typically assess the generated text against reference answers and the predicted trajectory against a recorded future \citep{sima2024drivelm,tian2024drivevlm}. High scores on these two outputs do not establish either that the trajectory depends on the content of the generated \chan{}, or that the \chan{} remains grounded in the visual evidence that it cites. A model can produce a plausible rationale while its trajectory follows shortcut priors, and it can receive strong scores under both separate text and trajectory evaluations. We therefore test two requirements implied by treating generated \chan{} as an explanation of a driving action:

\begin{itemize}
\itemsep0.2em
\item \emph{Does the trajectory of a driving VLA depend on the semantic content of the \chan{} it generates, or merely accompany it?}
\item \emph{When the world a driving VLA sees is changed, does its generated \chan{} and trajectory respond to this change?}
\end{itemize}

We answer these questions with two counterfactual interventions. For the first question, we keep the visual scene fixed, edit the model's own
generated \chan{}, and decode the resulting trajectory under the same inference configuration. For the second, we remove a model-identified action-relevant object from the camera frames and rerun inference, allowing the model to generate a new \chan{} and trajectory from the edited visual input. The
\chan{}-side intervention tests whether the stated reasoning steers the action. The image-side intervention tests whether this object-level change in visual evidence affects the model's generated \chan{} and its trajectory. Within this image-side comparison, observing the regenerated \chan{} and trajectory together reveals whether a revision to the model's stated reason is accompanied by a corresponding trajectory response, separating models that update both reasoning and action from those that update only one output.

Two calibrations make intervention responses comparable across models. Each \chan{}-side edit is paired within scene with a semantic-control paraphrase, so
a model that reacts to any token-level change is not mistaken for a model that reacts to changed meaning. We additionally report each intervention response relative to the model's response to camera blanking, which provides a model-specific reference scale for visual sensitivity. Applied to four driving VLAs on two benchmarks (\S\ref{sec:setup}), our thorough evaluations return a consistent picture of a weak coupling between a driving VLA's reasoning and action while also uncovering different model-specific shortcomings in their explainability.

\begin{table*}[t]
\centering
\small
\setlength{\tabcolsep}{4pt}
\renewcommand{\arraystretch}{1.15}
\begin{tabular}{p{0.10\linewidth} p{0.17\linewidth} p{0.17\linewidth} p{0.46\linewidth}}
\toprule
\textbf{Perturbation} & \textbf{Edit} & \textbf{Isolates} & \textbf{Example} \\
\midrule
\filler{}
& Contentless tokens, length-matched
& Whether any text is needed
& Keep lane to continue driving since no critical agent needs attention.
  $\rightarrow$
  .\textbackslash n.\textbackslash n.\textbackslash n\ldots,
  matched to the same token count. \\
\addlinespace[3pt]
\mistakeop{}
& LLM rewrite of the decision sentence
& Decision content, unconstrained rewrite
& Nudge left due to the stopped truck blocking the right side.
  $\rightarrow$
  Nudge right due to the stopped truck blocking the left side. \\
\addlinespace[3pt]
\negateop{}
& Template negation of the decision sentence
& Decision content, template-constrained
& Stop to yield to the pedestrian in the crosswalk.
  $\rightarrow$
  The path ahead is clear, so there is no need to stop;
  the ego vehicle should proceed at its current speed. \\
\addlinespace[3pt]
\swapop{}
& Graft another scene's decision sentence
& Decision content, real donor text
& Slow down due to upcoming speed bumps.
  $\rightarrow$
  Accelerate to maintain following distance to the lead vehicle. \\
\addlinespace[3pt]
\deleteop{}
& Remove the text
& Text presence
& Resume speed since the toll gate signal turns green.
  $\rightarrow$ empty string. \\
\bottomrule
\end{tabular}
\caption{\Chan{}-side text perturbations. Each content-changing perturbation (\mistakeop{}, \negateop{}, \swapop{}) is compared against a meaning-preserving \paraphrase{} of the same \chan{}.}
\label{tab:operators}
\end{table*}

\section{Related Work}
\label{sec:related}

\paragraph{\Chan{} in driving VLAs.}
Learned autonomous-driving systems span modular perception--prediction--planning stacks, end-to-end vision--action policies, and more recent driving vision--language--action models that combine multimodal language processing with trajectory or control prediction \citep{zhou2024vlmad,jiang2025vlasurvey,hu2025vla}. Within this progression, driving VLAs use vision--language representations either to provide scene or planning guidance to a separate driving policy \citep{tian2024drivevlm,jiang2024senna} or to jointly generate language and planning outputs in an end-to-end system \citep{hwang2024emma}. Language in these systems initially appeared as scene descriptions and question--answer outputs \citep{xu2023drivegpt4,sima2024drivelm}, while more recent models generate explicit deliberation through hierarchical scene analysis and meta-actions \citep{tian2024drivevlm}, planning-oriented question answering \citep{jiang2024senna,hwang2024emma}, adaptive chain-of-thought before action tokens \citep{zhou2025autovla,yuan2025autodrive,chi2025impromptu}, or single-sentence chains of causation associated with trajectory generation \citep{wang2025alpamayo}. We use \chan{} to refer to this scene-conditioned language generated during inference before action prediction. Although driving VLAs commonly present \chan{} as interpretability or explainable decision-making \citep{xu2023drivegpt4,tian2024drivevlm,zhou2025autovla,wang2025alpamayo}, existing driving evaluations do not directly test whether the semantic content of the free-text \chan{} truly explains the resulting trajectory. We therefore evaluate VLAs that generate scene-specific \chan{} and intervene on its reasoning while holding the remaining model inputs and decoding configuration fixed (\S\ref{sec:framework-word}).

\paragraph{Faithfulness evaluation for generated reasoning.}
Faithfulness research distinguishes explanations that reflect computation relevant to a model's output from explanations that are merely plausible to a human reader \citep{jacovi-goldberg-2020-towards,barez2025chain}. Behavioral faithfulness tests commonly intervene on either the reasoning trace or the input evidence. Reasoning-side interventions alter a generated reasoning trace or its intermediate steps and test whether the resulting model output depends on the edited content \citep{lanham2023measuring,paul2024making,sun2026facte}. Input-side interventions alter a cue or evidence available to the model and test whether the resulting explanation reflects the change \citep{turpin2023language,chen2025reasoning,atanasova-etal-2023-faithfulness}. These approaches capture different aspects of faithfulness, and recent work cautions that failure to verbalize an influential factor or success on a single metric is not a complete verdict on whether reasoning is faithful \citep{zaman2026really,young2026measuring}. Many prior behavioral tests read
their result from whether a discrete answer changes or remains unchanged, which is a natural evaluation endpoint for discrete-answer NLP tasks. In our work, we adapt intervention-based faithfulness evaluation to driving trajectories by measuring semantic \chan{}-to-action effects and the response of regenerated reasoning and trajectory to removal of a reasoning-selected visual object.

\section{Measuring the Reasoning--Action Coupling}
\label{sec:framework}

\fwsub{Motivation}
\label{sec:framework-motivation}

The closest prior study of the coupling between reasoning and action in driving models examines the question through training-time information ablations. \citet{song2025more} build a chain-of-thought driving corpus, train their own driving models on it with supervised fine-tuning and reinforcement learning, and compare models trained with and without chain-of-thought, visual input, and textual priors. Based on their finding that planning can remain strong without generated reasoning or visual input but degrades when textual priors are removed, they propose the Reasoning--Planning Decoupling Hypothesis, under which generated \chan{} serves as a `` byproduct rather than a causal mediator''. 

Although their ablations test whether training a model to generate reasoning improves planning, they do not directly intervene on the semantic content of reasoning produced by the same fixed model for the same scene. Consequently, they cannot determine whether that model's trajectory causally depends on what its generated \chan{} says. We instead evaluate state-of-the-art driving VLAs, that use different approaches to eliciting and structuring reasoning before action generation, and intervene directly on the generated \chan{} to test whether a change in meaning produces a corresponding change in the resulting trajectory. Our image-side counterfactuals complement this test by removing an action-relevant object from the camera input and examining whether its absence revises the generated \chan{} and whether the trajectory updates alongside that revised \chan{}.

\begin{figure*}[t]
\centering

\newlength{\humanstudycontentheight}
\setlength{\humanstudycontentheight}{3.15cm}

\begin{minipage}[t]{0.25\textwidth}
\vspace{0pt}%
\raggedright

\scriptsize
\setlength{\tabcolsep}{2.0pt}
\renewcommand{\arraystretch}{1.12}

\begin{minipage}[t][\humanstudycontentheight][t]{\linewidth}
\centering

\begin{tabular}{@{}lccc@{}}
\toprule
& \shortstack[c]{DiffuEraser\\\citep{li2025diffueraser}}
& \shortstack[c]{RORem\\\citep{li2025rorem}}
& \shortstack[c]{ObjectClear\\\citep{zhao2026objectclear}} \\
\midrule

Q1
& 3.19
& 3.40
& \textbf{4.45} \\

Q2
& 3.24
& 3.65
& \textbf{4.64} \\

Q3
& 3.21
& 3.44
& \textbf{4.67} \\

Q4
& 3.96
& 4.12
& \textbf{4.70} \\

\hline

Avg
& 3.40
& 3.65
& \textbf{4.61} \\

\bottomrule
\end{tabular}

\vfill
\end{minipage}

\captionsetup{
  type=table,
  font=small,
  labelfont=bf,
  justification=raggedright,
  singlelinecheck=false,
  skip=7pt
}
\captionof{table}{
Human evaluation of candidate object-removal methods, ratings between 1-5.
}
\label{tab:removal-human-study}
\end{minipage}\hspace{0.01\textwidth}%
\begin{minipage}[t]{0.73\textwidth}
\centering

\begin{minipage}[t][\dimexpr\humanstudycontentheight+10pt\relax][t]{\linewidth}
\vspace{0pt}%

\centering

\includegraphics[width=0.99\linewidth]{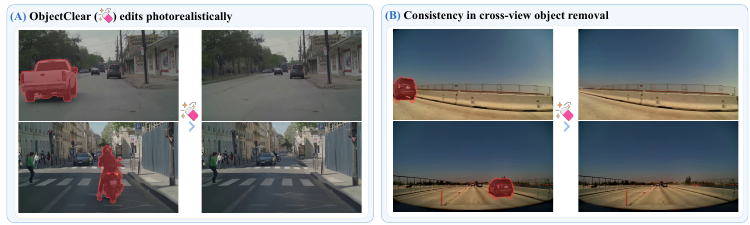}
\end{minipage}

\captionsetup{
  type=figure,
  hypcap=false,
  font=small,
  labelfont=bf,
  justification=raggedright,
  singlelinecheck=false,
  skip=7pt
}
\captionof{figure}{
ObjectClear \eraser{}-edited counterfactuals. The figures
illustrate removal quality, realism, visual artifacts, and preservation of
non-target scene content.
}
\label{fig:removal-qualitative}
\end{minipage}

\end{figure*}

\fwsub{Intervening on the generated \chan{}}
\label{sec:framework-word}

A scene $s$ supplies context $x = (I, e, g)$, consisting of camera frames $I$, ego-kinematic history $e$, and textual priors $g$. The model generates
\chan{} $c \sim p_\theta(\cdot \mid x)$ and then a trajectory $a \sim p_\theta(a \mid c, x)$ with waypoints $a = (w_1,\dots,w_H)$. For every driving VLA that we evaluate, trajectory decoding is conditioned on the preceding generated \chan{} and the scene context. Our interventions therefore operate on these two sources of conditioning: the \chan{} side replaces $c$, whereas the image side edits the camera frames $I$ within $x$.

On the \chan{} side, we feed edited versions of the model's own generated text and decode the trajectory under otherwise identical inference conditions. Let $\mathcal{D}(a, a')$ denote a trajectory-displacement metric between two decoded trajectories. Rather than compare an edited trajectory with ground truth, we measure its change relative to the trajectory decoded from the model's own unedited \chan{}, denoted $a_{\mathrm{self}}$. The raw response to a perturbation $T$ is 

\begin{equation} 
d_T^{\mathcal{D}}(s) = \mathcal{D}\!\left(a_T(s),\, a_{\mathrm{self}}(s)\right).
\end{equation}

This response does not by itself isolate content dependence, because a teacher-forced decoder can react to any token-level change, whether or not the change is meaningful. To account for this non-semantic wording sensitivity, we use a paired contrast that subtracts, for each scene, the response to a meaning-preserving \paraphrase,

\begin{equation}
\label{eq:paired}
\Delta_T^{\mathcal{D}}(s)
=
d_T^{\mathcal{D}}(s)
-
d_{\paraphrase{}}^{\mathcal{D}}(s).
\end{equation}

For each metric defined in Section~\ref{sec:protocol_and_metrics}, we report the mean paired effect $\Delta_T^{\mathcal{D}}(s)$ unless stated otherwise. Thus, $\Delta_T^{\mathcal{D}}(s)$ measures how much more the trajectory changes under perturbation $T$ than under a semantic-control rewording of the same \chan{}. Figure~\ref{fig:motivation} summarizes the \chan{}-side perturbations, and Table~\ref{tab:operators} lists each perturbation with an example. We next describe the construction and validation of the \paraphrase{} control, and the decision-classification procedure used to determine perturbation eligibility, verify the intended decision change under \mistakeop{} and \negateop{}, and select substitute reasoning sentences for \swapop{}.

\paragraph{Paraphrasing and decision classification} We use two auxiliary LLMs to construct the \chan{}-side text conditions and classify their decision content. Motivated by prior work on LLM-guided counterfactual generation \citep{fizle}, we use Llama-3.1-8B-Instruct \citep{llama31} to construct controlled text edits of the generated \chan{}. For the paraphrase control, Llama uses few-shot prompting with exemplar sentences from other scenes of the same subject-model and benchmark pair. Llama also produces the decision rewrite used by \mistakeop{}. We use a separate model from a different family, Qwen3.5-9B-Instruct \cite{qwen35}, to classify the generated \chan{} along independent longitudinal (\texttt{stop}, \texttt{decel}, \texttt{keep\_speed},
\texttt{accel}) and lateral (\texttt{left}, \texttt{straight}, \texttt{right}) axes. For each scene, these labels define a counterfactual decision target that differs from the model's original decision in its longitudinal direction, lateral direction, or both. This decision is introduced using one of three perturbations: a Llama rewrite in \mistakeop{}, a fixed template in \negateop{}, or a donor decision sentence in \swapop{}. Both auxiliary LLMs are deployed locally with vLLM.

\fwsub{Intervening on the images}
\label{sec:framework-images}

For each eligible scene $s$, let $o_s$ denote the object that the model's generated \chan{} identifies as action-relevant. We replace the factual camera input $I$ with a counterfactual input $I^{-o_s}$ in which $o_s$ is removed, while retaining the ego-kinematic history $e$ and textual priors $g$. This gives the counterfactual context $x^{-o_s} = (I^{-o_s}, e, g)$. Unlike the \chan{}-side intervention, we do not inject an edited \chan{} at the trajectory-decoding stage. Instead, we rerun the model freely on the edited visual input to generate both a new \chan{} and a new trajectory,

\begin{equation}
c^{-o_s} \sim p_\theta(\cdot \mid x^{-o_s}),
\qquad
a^{-o_s} \sim p_\theta(a \mid c^{-o_s}, x^{-o_s}).
\end{equation}

We compare the counterfactual outputs $(c^{-o_s}, a^{-o_s})$ with the factual outputs $(c, a)$ to determine whether removing a visual cause identified by the model revises its generated \chan{}, its trajectory, or both. For a trajectory-displacement metric $\mathcal{D}$, the image-side action response is

\begin{equation}
d_{-o_s}^{\mathcal{D}}(s)
=
\mathcal{D}\!\left(a^{-o_s}(s),\, a(s)\right).
\end{equation}

We interpret this trajectory response together with the regenerated \chan{} to determine whether a revision to the stated reason is accompanied by a corresponding action response. We report image-side trajectory responses  relative to the response induced by camera blanking, which provides a model specific reference scale for visual sensitivity. Section~\ref{sec:construction} describes the construction, validation, and eligibility criteria for the object-removal counterfactuals.

\section{Constructing the Visual Counterfactuals}
\label{sec:construction}

To implement image-side interventions, we construct counterfactual frames where a specific object is removed while the remaining scene is preserved as closely as possible. Our inspection of the DriveLM \cite{sima2024drivelm} dataset annotations revealed that most samples do not provide a single, removable "key" object that the VLA's own reasoning will identify as directly relevant to its predicted maneuver. We therefore build object-removal counterfactuals on PhysicalAI-AV \citep{nvidia2025physicalai} dataset based on the key objects described in the generated reasoning of each evaluated VLA on this dataset (\alponeS{}, \alprS{}, \autovlaS{}). The selected key objects are model-specific because each VLA consumes a different temporal context and may verbalize different key objects in its reasoning even for the same driving scene. The following subsections describe how we localize and track the key object, and how we select the most reliable object-removal method through human evaluation before using it to construct the visual counterfactuals.

\dssub{Locating the key object}
\label{sec:construction-mask}

The model-generated reasoning identifies the key object in language, but the counterfactual editor requires a spatial mask for that object in every relevant input frame. To reduce the manual annotation required for this process, we first parse each model's generated reasoning into a key-object phrase and determine whether that phrase refers to a single removable physical object. We use the extracted phrase to prompt an open-vocabulary detector \citep{liu2024groundingdino} on the first frame in that camera stream in which the key object is visible. Each resulting box prompts a segmentation model \citep{kirillov2023sam}, and we propagate the resulting mask through the remaining frames of that camera stream using a video segmentation model \citep{ravi2025sam2}. Additionally, the selected key object may be visible in multiple camera views (\autovlaS{} uses three camera views, whereas \alponeS{} and \alprS{} use four). We therefore perform view-specific grounding, segmentation, and temporal mask propagation for every camera stream in which the object is visible. Automatic masks are then reviewed and corrected by manual effort through an interactive tool. The retained view- and time-indexed masks provide the fixed object support used by every candidate removal method in the following subsection.

\dssub{Selecting an object-removal method for counterfactual generation}
\label{sec:construction-edit}

Object-removal models reconstruct the background that would plausibly occupy a user-specified object region. Recent diffusion-based methods have improved photorealistic object removal by generating plausible background content inside a supplied mask \citep{li2025rorem,zhao2026objectclear,jiang2025smarteraser,sun2025attentiveeraser}. Related video-inpainting methods additionally model temporal consistency so that the reconstructed content remains stable across adjacent frames \citep{zhou2023propainter,li2025diffueraser}. To generate the visual counterfactuals we need, we compare three masked removal methods. RORem \citep{li2025rorem} and ObjectClear \citep{zhao2026objectclear} are image-based object-removal methods, whereas DiffuEraser \citep{li2025diffueraser} is a video-inpainting method. ObjectClear is specifically interesting because it can adapt the edited region beyond the input object mask to remove associated effects of the target, including cast shadows, reflections, and glare. This capability is useful for our counterfactuals because a key object can remain visually implied by effects outside its segmentation mask.

We apply the three candidate removal methods to a stratified random sample of 100 frame--method outputs, with strata defined by camera view and key-object size. Five annotators rate 100 presented clips each on a five-point scale using four questions. Q1 measures overall realism without revealing the original scene, asked when annotators first view only the edited frame and rate it by itself. Annotators then view the original and edited frames side by side to rate Q2--Q4, which assess target removal and reconstruction quality, elimination of visual effects associated with the key object, and preservation of regions outside the edited area respectively.  Table~\ref{tab:removal-human-study} reports the mean ratings. ObjectClear receives the highest rating on every question with an overall mean of $4.61$, reasonably due to its advantage is in removing visual effects associated with the key object, including shadows, reflections, glare, and local shading while maintaining high quality of preservation. We therefore use ObjectClear, with 20 diffusion steps and a guidance scale of 2.5, to construct the final counterfactual inputs, and examples are shown in Figure~\ref{fig:removal-qualitative}.

Based on this framework, we generate counterfactual frames for model--scene pairs whose candidate masks pass key-object localization and manual mask review. We then perform a final post-edit quality check to verify that the key object and its visually salient associated effects are absent from the generated frames. We exclude a model--scene pair when the object remains visible, when the edit leaves evidence that continues to identify the object, or when the edited frames contain a substantial reconstruction artifact (e.g., a removed vehicle when its red taillight glare remains visible after the vehicle itself has been erased). We generate photorealistic object-removal inputs from \textbf{1,332 frame-level masks} associated with 157 \alponeS{} key-object instances, \textbf{1,403 frame-level masks} associated with 164 \alprS{} instances, and \textbf{511 frame-level masks} associated with 104 \autovlaS{} instances.

\section{Experimental Setup}
\label{sec:setup}

\begin{table*}[t]
\centering
\footnotesize

\setlength{\tabcolsep}{2.3pt}

\renewcommand{\arraystretch}{1.18}

\begin{adjustbox}{max width=\textwidth}
\begin{tabular}{@{}l|c|c|c|c|c|c@{}}
\toprule

\multirow{2}{*}{\textbf{Dataset/Model}}
& \multicolumn{1}{|c}{\textbf{\deleteop{}}}
& \multicolumn{1}{c}{\textbf{\filler{}}}
& \multicolumn{1}{c}{\textbf{\swapop{}}}
& \multicolumn{1}{c}{\textbf{\mistakeop{}}}
& \multicolumn{1}{c}{\textbf{\negateop{}}}
& \multicolumn{1}{|c}{\textbf{$C_{\text{cam}}$}} \\

\cline{2-7}

& \textsc{ADE/FDE}
& \textsc{ADE/FDE}
& \textsc{ADE/FDE}
& \textsc{ADE/FDE}
& \textsc{ADE/FDE}
& \textsc{ADE/FDE} \\

\midrule

\multicolumn{7}{@{}l@{}}{\scriptsize\itshape DriveLM} \\[-0.80ex]

\autovlaS{}
& \afpair{\mathbf{+0.450}}{\mathbf{+0.977}}
& \afpair{\underline{+0.393}}{\underline{+0.841}}
& \afpair{+0.012}{+0.035}
& \afpair{+0.196}{+0.394}
& \afpair{+0.016}{+0.032}
& \afpair{2.37}{5.78} \\

\orionM{}
& \afpair{\underline{+0.178}}{\underline{+0.277}}
& \afpair{\mathbf{+0.194}}{\mathbf{+0.300}}
& \afpair{+0.022}{+0.034}
& \afpair{-0.018}{-0.026}
& \afpair{+0.107}{+0.150}
& \afpair{2.41}{4.92} \\[-0.10ex]

\multicolumn{1}{@{}l@{}}{\scriptsize\itshape PhysicalAI-AV}
& \multicolumn{6}{@{}l@{}}{%
  \rlap{\raisebox{0.35ex}{\rule{0.667\textwidth}{0.4pt}}}%
} \\[0.20ex]

\autovlaS{}
& \afpair{\mathbf{+0.667}}{\mathbf{+1.542}}
& \afpair{\underline{+0.440}}{\underline{+0.984}}
& \afpair{+0.112}{+0.264}
& \afpair{+0.261}{+0.592}
& \afpair{+0.037}{+0.216}
& \afpair{2.58}{6.08} \\

\alprS{}
& \afpair{\underline{+0.219}}{\underline{+0.555}}
& \afpair{+0.027}{+0.094}
& \afpair{+0.070}{+0.211}
& \afpair{\mathbf{+0.483}}{\mathbf{+1.285}}
& \afpair{+0.101}{+0.294}
& \afpair{3.24}{9.44} \\

\alponeS{}
& \afpair{+0.005}{+0.005}
& \afpair{+0.098}{+0.294}
& \afpair{+0.475}{+1.649}
& \afpair{\mathbf{+0.639}}{\mathbf{+2.093}}
& \afpair{\underline{+0.600}}{\underline{+1.971}}
& \afpair{3.28}{9.26} \\

\bottomrule
\end{tabular}
\end{adjustbox}

\caption{Results of \chan{} side perturbations on the DriveLM and
PhysicalAI-AV validation sets. Each entry reports mean ADE / FDE trajectory changes. The final column gives the camera-side per-model normalization numbers.}
\label{tab:wordside-combined-mean}
\end{table*}

\dssub{Models}
\label{sec:models}

Recent driving VLAs increasingly generate scene-conditioned reasoning alongside their predicted trajectories, motivated in part by the prospect of making
driving decisions more explainable \citep{hu2025vla}. We evaluate four such models whose reasoning reaches the action prediction pathway through distinct
interfaces. \autovlaS{} \citep{zhou2025autovla} couples a Qwen2.5-VL backbone to a codebook of physical action tokens, so interventions on the generated \chan{} directly alter the autoregressive context from which waypoint tokens are decoded. \alponeS{} \citep{nvidia2026alpamayo1.5} and \alprS{} \citep{wang2025alpamayo} pair a Cosmos-Reason backbone, which generates a single-sentence chain of causation, with a diffusion action expert conditioned on that backbone, and differ only in post-training: \alprS{} is trained with supervised fine-tuning (SFT) only, whereas \alponeS{} additionally receives reinforcement-learning post-training, providing a training-regime comparison within a shared architecture. \orionM{} \citep{fu2025orion} aggregates multi-view history with a query-based temporal module, reasons in a language model, and conditions a generative planner on a planning-token embedding rather than on the emitted text. The generated \chan{} differs in form across models, ranging from the single-sentence chain of causation used by the Alpamayo models to AutoVLA's structured chain-of-thought and ORION's multi-round QA transcript. The decision-targeted \mistakeop{}, \negateop{}, and \swapop{} interventions edit the full chain of causation for Alpamayo, the stated action rationale block for AutoVLA, and the planning justification following ORION's action recommendation. In addition, evaluated models differ not only in their action decoders, but also in the representation through which reasoning-related information reaches planning. This variation lets us distinguish findings tied to one interface from those that recur across several ways of connecting reasoning and action. We also examined the feasibility of evaluating additional driving VLAs (OmniDrive \citep{wang2025omnidrive}, Impromptu \citep{chi2025impromptu}, and AutoDrive-R2 \citep{yuan2025autodrive}), but eventually exclude them because their reasoning--action interfaces do not support a localized semantic intervention on the generated reasoning. OmniDrive predicts trajectories with a planning head that is not conditioned on the reasoning output. Impromptu produces a fixed planning template rather than scene-specific decision content that can be semantically edited. AutoDrive-R2 emits coordinate trajectories through free-text output, without a localized commitment statement that can be edited without disrupting coordinate decoding. All four evaluated models are run locally from their released checkpoints without any fine-tuning.

\begin{figure*}[t]
  \centering
  \includegraphics[width=\textwidth]{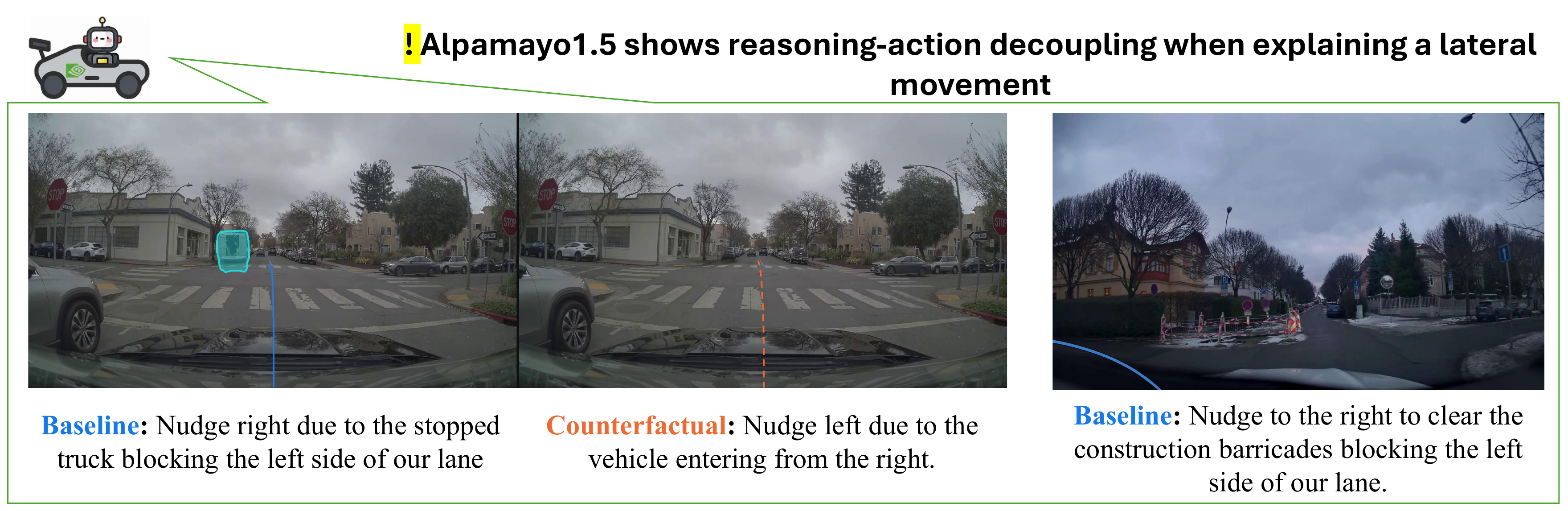}
  \caption{
    \alponeS{}'s reasoning-action failure cases: \textbf{(Left)} Though the image counterfactual test leads to a change in reasoning, the generated trajectories stay approximately the same. \textbf{(Right)} Investigating \alponeS{}'s generated reasoning on PhysicalAI-AV events shows cases where the model's Chain-of-Causation does not reflect its decision-making process leading up to an action.
  }
  \label{fig:alpamayo}
\end{figure*}

\dssub{Datasets}
\label{sec:datasets}

We evaluate the models on fixed, held-out evaluation sets of two mainstream driving datasets. Restricting the study to these partitions follows the datasets' native evaluation protocols and evaluates the intervention procedure on scenes designated for evaluation rather than model fitting. The DriveLM validation subset contains 799 scenes with graph-structured language annotations over nuScenes \citep{sima2024drivelm,caesar2020nuscenes} that  provide a structured description of the driving context (key objects, driving decisions, etc.) associated with each sample. We use this subset to evaluate \autovlaS{} and \orionM{}, both of which operate on nuScenes-derived driving inputs and produce scene-conditioned language alongside driving actions. The PhysicalAI-AV validation subset contains 349 real-world driving events with synchronized camera observations and ego trajectories \citep{nvidia2025physicalai}. We evaluate \alponeS{} and \alprS{} on this subset because it provides the native input format for the Alpamayo models. We also evaluate \autovlaS{} here, allowing us to test whether reasoning-side intervention pattern persists across different evaluation distributions. To construct AutoVLA inputs on PhysicalAI-AV, we derive driving instructions following the AutoVLA protocol used for nuScenes and use the provided trajectory and camera data, sampled at 2\,Hz in order to form its four-frame input at the temporal spacing used during native AutoVLA training.

\dssub{Protocol and metrics}
\label{sec:protocol_and_metrics}

Results are based on three-seed runs (seeds 42/43/44) unless stated otherwise. Comparisons are paired within both scene and seed, so each counterfactual  trajectory is evaluated against a baseline trajectory generated from the same scene and seed. Before evaluating a \chan{}-side intervention, we check the text-injection procedure by decoding the model's own unedited \chan{}, re-injecting it through the same intervention path, and regenerating the trajectory using the same seed and decoding configuration. This check ensures that the semantic-edit and paraphrase conditions are evaluated through the same injection interface. Any variation associated with that shared interface is therefore not attributed to reasoning content, since every reported \chan{}-side effect is the paired difference between a semantic edit and a meaning-preserving \paraphrase{} control that traverses the same path. For image-side interventions, factual and counterfactual frames are processed through the same evaluation pipeline, with identical camera ordering, timestamps, resizing, normalization, ego-state inputs, driving instructions, seeds, and decoding configurations. 

We measure intervention-induced trajectory change using Average Displacement Error (ADE) and Final Displacement Error (FDE). ADE measures the average
distance between corresponding waypoints in two predicted trajectories over the full prediction horizon, while FDE measures the distance between their final waypoints. ADE therefore captures how much an intervention changes the trajectory overall, whereas FDE captures how much it changes the predicted
endpoint. Although ADE and FDE conventionally compare a predicted trajectory with an expert trajectory, we apply them here to baseline/factual and intervened/counterfactual trajectories from the same model. They therefore measure intervention-induced trajectory change rather than planning error.

\section{Findings}
\label{sec:findings}

Subsection headings state the findings, and each subsection reports the experiment that was run, the observed pattern, the reference condition or criterion used to interpret it, and the insight specific to that section. By default, all experiments follow the common protocol described
in Section~\ref{sec:protocol_and_metrics}.  We summarize the main patterns using ADE because ADE and FDE yield the same qualitative ordering of models and interventions throughout our results. A larger value means that an intervention moved the model's predicted action more compared to the matched reference, not that the resulting trajectory is necessarily better, worse, or less safe. Unless otherwise noted, percentages  express the same trajectory change relative to the model's camera-blanking response. For example, a $20\%$ effect means that the text-side intervention moves the trajectory by one fifth of the change induced by removing camera input, which is used as a within-model reference scale.

\begin{figure*}[t]
  \centering
  \includegraphics[width=\textwidth]{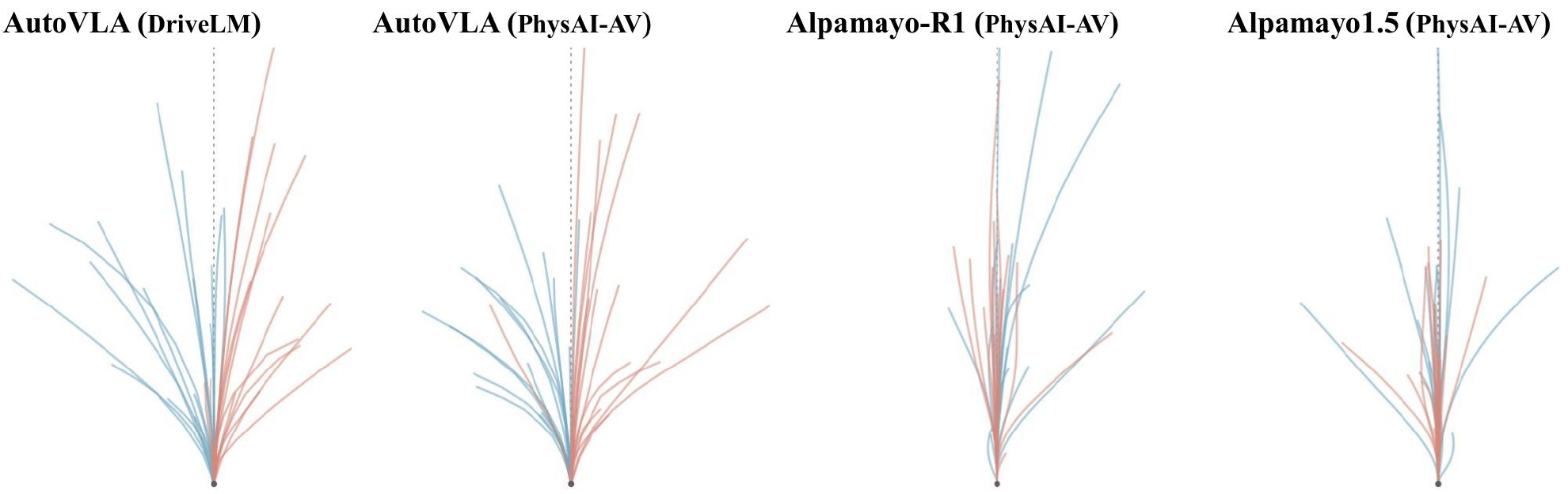}
  \caption{
    \textbf{Factual trajectories grouped by stated lateral direction.}
    Light blue trajectories correspond to samples whose generated \chan{} states
    \textcolor{blue!55!black}{left}; light red trajectories correspond to samples
    whose \chan{} states \textcolor{red!65!black}{right}. Each panel shows factual
    trajectories decoded from the corresponding model and evaluation set.
  }
  \label{fig:directional-visualization}
\end{figure*}

\subsection{Models split between \chan{}-presence dependence and content-sensitive action}
\label{sec:f-mediation}

A trajectory change after one \chan{} intervention does not by itself show that the model acts on reasoning content. We therefore compare several
content-changing edits with \deleteop{} and \filler{} controls, which test whether an apparent effect instead arises from the presence or surface form of the \chan{} input. We ask how consistently and how strongly each model responds to content changes after controlling for paraphrase-induced wording variation.

\expruni For every model, we teacher-force the full operator battery from Table~\ref{tab:operators} and compute paired effects against the meaning-preserving \paraphrase{} null in Eq.~\ref{eq:paired}. This comparison asks whether a trajectory change is attributable to the driving content of the generated \chan{}, rather than to passing modified text through the intervention interface.

\resruni Interestingly, the models' resulting actions as seen in Table~\ref{tab:wordside-combined-mean} are not uniform across all driving VLAs. Firstly, \autovlaS{} and \orionM{} respond primarily to the availability or surface form of the \chan{} input. Compared to the content-changing \swapop{}, \mistakeop{}, and \negateop{} interventions, they respond to \deleteop{} and \filler{} \textbf{at much larger magnitudes} ($15.65\%$ vs. $3.26\%$ ADE, averaging both models). \mistakeop{} for ORION, which tells the model to go in an opposite direction, is indistinguishable from the \paraphrase{} null. Thus, for \autovlaS{} and \orionM{}, the fact that text occupies the \chan{} slot matters more than the counterfactual driving content carried by that text in terms of steering their actions. \alprS{}, the earlier version of the Alpamayo models, responds most strongly to \mistakeop{}, although \deleteop{} produces its second-largest effect. Additionally, this content sensitivity does not extend to the other content-changing perturbations since \swapop{} and \negateop{} remain below $3.20\%$ ADE. On the other hand, \alponeS{} shows
a consistent content-sensitive pattern. All three content-changing interventions \textbf{produce substantial effects}, reaching $14.48\%$--$19.46\%$ of
the camera-side ADE response. However, it is important to note that even the decision rewrite in \mistakeop{}, which produces the largest effect, reaches only $19.46\%$ of the camera-side ADE response.

\subsection{Object removal updates reasoning and action, but their updates can disagree}
\label{sec:f-image-side}

The reasoning-side interventions in Section~\ref{sec:f-mediation} test whether the semantic content of generated \chan{} steers the predicted trajectory while the visual scene remains fixed. They do not ask the complementary question of whether the generated \chan{} is actually grounded in visual evidence when the scene changes, which we test through running these VLAs on object-removal counterfactuals. However, simply "demonstrating grounding" by updating both reasoning and action is not sufficient to establish that the trajectory follows the regenerated reasoning. We therefore examine both whether object removal changes the generated reasoning and trajectory, and whether the resulting trajectory is consistent with the maneuver described in the regenerated reasoning.

\expruni With the validated object-removal counterfactuals constructed in Section~\ref{sec:construction}, we rerun \autovlaS{}, \alprS{}, and \alponeS{} on PhysAI-AV scenes with object-removed inputs. The models regenerate both their \chan{} and trajectories from the edited camera input. For every counterfactual output, we manually inspect whether the regenerated \chan{} continues to mention the removed object or treats it as still present and action-relevant. We also exhaustively inspect the factual and counterfactual trajectories alongside their corresponding generated \chan{} to assess whether the observed action change is consistent with the revised stated maneuver.

\resruni Across the evaluated object-removal counterfactuals, the regenerated \chan{} does not continue to cite the removed object as present or action-relevant.The regenerated reasoning consistently drops the removed object from its account of the scene, providing encouraging evidence that its stated object references update with the edited visual evidence. Despite this update, our inspection shows that the counterfactual trajectory does not always implement the maneuver stated in the regenerated \chan{}. As shown by the qualitative example in Figure~\ref{fig:alpamayo}, the reasoning can be updated after object removal while the corresponding trajectory does not stay faithful to it. These cases show that joint responsiveness to changed visual evidence does not by itself guarantee reasoning--action alignment. 

\subsection{A directional finding: stated direction in reasoning does not reliably predict trajectories}
\label{sec:f-direction}

The preceding interventions test whether changing generated \chan{} or visual
evidence changes the predicted trajectory. However, neither intervention
directly determines whether the unedited \chan{} accurately describes the
model's own factual action. Assessing whether a driving VLA's generated
reasoning describes its action is not straightforward through conventional
ground-truth evaluation. We instead probe this question through a simple
within-model comparison. Rather than asking whether the model drives correctly,
we ask whether its stated lateral decision is consistent with the action
decoded from that same output. This provides a directly measurable test of whether the generated \chan{} describes the model's
own factual trajectory.

\expruni We reuse the lateral-direction labels produced by Qwen for the
reasoning-side intervention procedure \cite{qwen35}. We retain factual scenes
in which the model's generated \chan{} commits to a leftward or rightward
maneuver, and exclude scenes labeled as straight because they do not make a
lateral directional claim. For each committed scene, we determine the side
reached by the decoded factual trajectory from its lateral endpoint. 
We then report the fraction of scenes for which the stated and decoded directions agree, which we call the directional
follow-through rate.

\resruni Directional follow-through sharply separates \autovlaS{} from the
Alpamayo models. On DriveLM, \autovlaS{} follows its stated leftward or
rightward decision on $187$ of $202$ committed scenes. Its left-stated and right-stated trajectories are separated by
$7.06$\,m at the median lateral endpoint. \autovlaS{} also reaches $89.2\%$
follow-through on PhysicalAI-AV ($66$ of $74$ committed scenes).
By contrast, \alponeS{} follows its stated direction on only $71$ of $155$
committed scenes, contradicting its own stated direction
on $84$ scenes. Its left-stated and right-stated trajectories differ by only
$0.01$\,m at the median endpoint. \alprS{} shows the same pattern, following
through on $59$ of $121$ committed scenes, with a median
left/right separation of $0.27$\,m. Thus, for both Alpamayo models, stating
that the vehicle will turn left rather than right provides almost no
information about which side their factual trajectories reach, as visualized in Figure~\ref{fig:directional-visualization}. \emph{Due to \orionM{}'s  small number of committed scenes, results cannot be concluded.}

\subsection{The reasoning--action disconnect is not a benchmark artifact}
\label{sec:f-transfer}

The results in Section~\ref{sec:f-mediation} identify worrying patterns on current models. However,
reasoning--action conclusions drawn from one evaluation set can be difficult to
separate from the particular scenes, maneuvers, and generated reasoning
instances available in that set. We therefore ask whether the reasoning--action signature observed for
one sample model, \autovlaS{}, is stable when the same model is evaluated on a second set of
driving events. Holding the model weights and intervention protocol fixed lets
us test whether changing the evaluation context restores sensitivity to
counterfactual reasoning content, or whether the same patterns recur.

\expruni We evaluate the same \autovlaS{} weights with the same \chan{}-side perturbation set and paired evaluation protocol on DriveLM and PhysicalAI-AV. Section~\ref{sec:datasets} describes the construction of \autovlaS{} inputs on PhysicalAI-AV, including the four-frame camera input and
driving instructions derived according to the native \autovlaS{} protocol.

\resruni \autovlaS{}'s pattern replicates on both DriveLM and PhysicalAI-AV, as \deleteop{} and \filler{} remain the two largest effects on both sets, while the content-changing edits are smaller. Although the absolute response
magnitudes and the order among the content-changing edits differ across the two sets, \deleteop{} and \filler{} dominate on both. This recurrence shows that across both evaluated sets,
changing whether or how the \chan{} is supplied matters more to \autovlaS{}'s
action than changing most of its stated driving content.

\section{Conclusion}
Driving VLAs present generated reasoning as an account of why the vehicle acts.
For this account to be reliable, the trajectory must depend on the reasoning's
semantic content, and both reasoning and action must remain responsive to
changes in the visual scene. Our thorough analysis tests these requirements
through paired reasoning-side edits, object-removal image counterfactuals, and
model-internal comparisons between stated and decoded driving directions.

Across four driving VLAs, we find that this reasoning--action relationship is
model-dependent and often weaker than the generated text suggests. For
\autovlaS{} and \orionM{}, trajectory changes are driven more by the presence
or surface form of the reasoning input than by its counterfactual driving
content. The Alpamayo models show stronger but different forms of content
sensitivity, while their stated leftward and rightward decisions do not
reliably describe the lateral directions of their factual trajectories.
Object removal further shows that though models can revise their generated reasoning
when visual evidence changes, we reveal that the updated
trajectory does not always implement the maneuver described by that reasoning.

These findings show that plausible, visually responsive reasoning is not by
itself a faithful explanation of a driving VLA's action. Future driving VLAs
should therefore be evaluated not only by reasoning quality and trajectory
accuracy separately, but also by whether the stated reasoning causally and
consistently accounts for the action it accompanies.

{
    \small
    \bibliographystyle{plainnat}
    \bibliography{refs}

\begin{thebibliography}{40}
\providecommand{\natexlab}[1]{#1}
\providecommand{\url}[1]{\texttt{#1}}
\expandafter\ifx\csname urlstyle\endcsname\relax
  \providecommand{\doi}[1]{doi: #1}\else
  \providecommand{\doi}{doi: \begingroup \urlstyle{rm}\Url}\fi

\bibitem[Atanasova et~al.(2023)Atanasova, Camburu, Lioma, Lukasiewicz,
  Simonsen, and Augenstein]{atanasova-etal-2023-faithfulness}
Pepa Atanasova, Oana-Maria Camburu, Christina Lioma, Thomas Lukasiewicz,
  Jakob~Grue Simonsen, and Isabelle Augenstein.
\newblock Faithfulness tests for natural language explanations.
\newblock In \emph{Proceedings of the 61st Annual Meeting of the Association
  for Computational Linguistics (Volume 2: Short Papers)}, pages 283--294,
  2023.

\bibitem[Barez et~al.(2025)Barez, Wu, Arcuschin, Lan, Wang, Siegel, Collignon,
  Neo, Lee, Paren, et~al.]{barez2025chain}
Fazl Barez, Tung-Yu Wu, Iv{\'a}n Arcuschin, Michael Lan, Vincent Wang, Noah
  Siegel, Nicolas Collignon, Clement Neo, Isabelle Lee, Alasdair Paren, et~al.
\newblock Chain-of-thought is not explainability.
\newblock Preprint, alphaXiv.
  \url{https://aigi.ox.ac.uk/publications/chain-of-thought-is-not-explainability/},
  2025.

\bibitem[Bhattacharjee et~al.(2024)Bhattacharjee, Moraffah, Garland, and
  Liu]{fizle}
Amrita Bhattacharjee, Raha Moraffah, Joshua Garland, and Huan Liu.
\newblock Zero-shot {LLM}-guided counterfactual generation: A case study on
  {NLP} model evaluation.
\newblock In \emph{2024 IEEE International Conference on Big Data}, pages
  1243--1248. IEEE, 2024.

\bibitem[Caesar et~al.(2020)Caesar, Bankiti, Lang, Vora, Liong, Xu, Krishnan,
  Pan, Baldan, and Beijbom]{caesar2020nuscenes}
Holger Caesar, Varun Bankiti, Alex~H. Lang, Sourabh Vora, Venice~Erin Liong,
  Qiang Xu, Anush Krishnan, Yu~Pan, Giancarlo Baldan, and Oscar Beijbom.
\newblock {nuScenes}: A multimodal dataset for autonomous driving.
\newblock In \emph{Proceedings of the IEEE/CVF Conference on Computer Vision
  and Pattern Recognition}, pages 11618--11628, 2020.

\bibitem[Chen et~al.(2025)Chen, Benton, Radhakrishnan, Uesato, Denison,
  Schulman, Somani, Hase, Wagner, Roger, et~al.]{chen2025reasoning}
Yanda Chen, Joe Benton, Ansh Radhakrishnan, Jonathan Uesato, Carson Denison,
  John Schulman, Arushi Somani, Peter Hase, Misha Wagner, Fabien Roger, et~al.
\newblock Reasoning models don't always say what they think.
\newblock \emph{arXiv preprint arXiv:2505.05410}, 2025.

\bibitem[Chi et~al.(2025)Chi, ang Gao, Liu, Liu, Liu, Li, Yang, Yu, Wang, Li,
  et~al.]{chi2025impromptu}
Haohan Chi, Huan ang Gao, Ziming Liu, Jianing Liu, Chenyu Liu, Jinwei Li,
  Kaisen Yang, Yangcheng Yu, Zeda Wang, Wenyi Li, et~al.
\newblock Impromptu {VLA}: Open weights and open data for driving
  vision-language-action models.
\newblock In \emph{Advances in Neural Information Processing Systems},
  volume~38, 2025.

\bibitem[Dubey et~al.(2024)Dubey, Jauhri, Pandey, Kadian, Al-Dahle, Letman,
  Mathur, Schelten, Yang, Fan, et~al.]{llama31}
Abhimanyu Dubey, Abhinav Jauhri, Abhinav Pandey, Abhishek Kadian, Ahmad
  Al-Dahle, Aiesha Letman, Akhil Mathur, Alan Schelten, Amy Yang, Angela Fan,
  et~al.
\newblock The {Llama} 3 herd of models.
\newblock \emph{arXiv preprint arXiv:2407.21783}, 2024.

\bibitem[Fu et~al.(2025)Fu, Zhang, Zhao, Cui, Liang, Zhang, Zhang, Xie, Wang,
  and Bai]{fu2025orion}
Haoyu Fu, Diankun Zhang, Zongchuang Zhao, Jianfeng Cui, Dingkang Liang, Chong
  Zhang, Dingyuan Zhang, Hongwei Xie, Bing Wang, and Xiang Bai.
\newblock {ORION}: A holistic end-to-end autonomous driving framework by
  vision-language instructed action generation.
\newblock In \emph{Proceedings of the IEEE/CVF International Conference on
  Computer Vision}, pages 24823--24834, 2025.

\bibitem[Hu et~al.(2025)Hu, Liu, Wang, Zhu, Liang, Kong, Zhao, Gong, Cen,
  Huang, et~al.]{hu2025vla}
Tianshuai Hu, Xiaolu Liu, Song Wang, Yiyao Zhu, Ao~Liang, Lingdong Kong,
  Guoyang Zhao, Zeying Gong, Jun Cen, Zhiyu Huang, et~al.
\newblock Vision-language-action models for autonomous driving: Past, present,
  and future.
\newblock \emph{arXiv preprint arXiv:2512.16760}, 2025.

\bibitem[Hwang et~al.(2025)Hwang, Xu, Lin, Hung, Ji, Choi, Huang, He,
  Covington, Sapp, et~al.]{hwang2024emma}
Jyh-Jing Hwang, Runsheng Xu, Hubert Lin, Wei-Chih Hung, Jingwei Ji, Kristy
  Choi, Di~Huang, Tong He, Paul Covington, Benjamin Sapp, et~al.
\newblock {EMMA}: End-to-end multimodal model for autonomous driving.
\newblock \emph{Transactions on Machine Learning Research}, 2025.

\bibitem[Jacovi and Goldberg(2020)]{jacovi-goldberg-2020-towards}
Alon Jacovi and Yoav Goldberg.
\newblock Towards faithfully interpretable {NLP} systems: How should we define
  and evaluate faithfulness?
\newblock In \emph{Proceedings of the 58th Annual Meeting of the Association
  for Computational Linguistics}, pages 4198--4205, 2020.

\bibitem[Jiang et~al.(2024)Jiang, Chen, Liao, Zhang, Yin, Zhang, Huang, Liu,
  and Wang]{jiang2024senna}
Bo~Jiang, Shaoyu Chen, Bencheng Liao, Xingyu Zhang, Wei Yin, Qian Zhang, Chang
  Huang, Wenyu Liu, and Xinggang Wang.
\newblock Senna: Bridging large vision-language models and end-to-end
  autonomous driving.
\newblock \emph{arXiv preprint arXiv:2410.22313}, 2024.

\bibitem[Jiang et~al.(2025{\natexlab{a}})Jiang, Zhang, Qi, Zhou, Zhang, and
  Li]{jiang2025smarteraser}
Longtao Jiang, Zhendong Zhang, Jinyang Qi, Yueting Zhou, Wengang Zhang, and
  Houqiang Li.
\newblock {SmartEraser}: Remove anything from images using masked-region
  guidance.
\newblock In \emph{Proceedings of the IEEE/CVF Conference on Computer Vision
  and Pattern Recognition}, 2025{\natexlab{a}}.

\bibitem[Jiang et~al.(2025{\natexlab{b}})Jiang, Huang, Qian, Luo, Zhu, Zhong,
  Tang, Kong, Wang, Jiao, et~al.]{jiang2025vlasurvey}
Sicong Jiang, Zilin Huang, Kangan Qian, Ziang Luo, Tianze Zhu, Yang Zhong,
  Yihong Tang, Menglin Kong, Yunlong Wang, Siwen Jiao, et~al.
\newblock A survey on vision-language-action models for autonomous driving.
\newblock \emph{arXiv preprint arXiv:2506.24044}, 2025{\natexlab{b}}.

\bibitem[Kirillov et~al.(2023)Kirillov, Mintun, Ravi, Mao, Rolland, Gustafson,
  Xiao, Whitehead, Berg, Lo, Doll{\'a}r, and Girshick]{kirillov2023sam}
Alexander Kirillov, Eric Mintun, Nikhila Ravi, Hanzi Mao, Chloe Rolland, Laura
  Gustafson, Tete Xiao, Spencer Whitehead, Alexander~C. Berg, Wan-Yen Lo, Piotr
  Doll{\'a}r, and Ross Girshick.
\newblock Segment anything.
\newblock In \emph{Proceedings of the IEEE/CVF International Conference on
  Computer Vision}, 2023.

\bibitem[Lanham et~al.(2023)Lanham, Chen, Radhakrishnan, Steiner, Denison,
  Hernandez, Li, Durmus, Hubinger, Kernion, et~al.]{lanham2023measuring}
Tamera Lanham, Anna Chen, Ansh Radhakrishnan, Benoit Steiner, Carson Denison,
  Danny Hernandez, Dustin Li, Esin Durmus, Evan Hubinger, Jackson Kernion,
  et~al.
\newblock Measuring faithfulness in chain-of-thought reasoning.
\newblock \emph{arXiv preprint arXiv:2307.13702}, 2023.

\bibitem[Li et~al.(2025{\natexlab{a}})Li, Yang, Guo, and Zhang]{li2025rorem}
Ruibin Li, Tao Yang, Song Guo, and Lei Zhang.
\newblock {RORem}: Training a robust object remover with human-in-the-loop.
\newblock In \emph{Proceedings of the IEEE/CVF Conference on Computer Vision
  and Pattern Recognition}, 2025{\natexlab{a}}.

\bibitem[Li et~al.(2025{\natexlab{b}})Li, Xue, Ren, and Bo]{li2025diffueraser}
Xiaowen Li, Haolan Xue, Peiran Ren, and Liefeng Bo.
\newblock {DiffuEraser}: A diffusion model for video inpainting.
\newblock \emph{arXiv preprint arXiv:2501.10018}, 2025{\natexlab{b}}.

\bibitem[Liu et~al.(2024)Liu, Zeng, Ren, Li, Zhang, Yang, Jiang, Li, Yang, Su,
  Zhu, and Zhang]{liu2024groundingdino}
Shilong Liu, Zhaoyang Zeng, Tianhe Ren, Feng Li, Hao Zhang, Jie Yang, Qing
  Jiang, Chunyuan Li, Jianwei Yang, Hang Su, Jun Zhu, and Lei Zhang.
\newblock Grounding {DINO}: Marrying {DINO} with grounded pre-training for
  open-set object detection.
\newblock In \emph{European Conference on Computer Vision}, 2024.

\bibitem[{NVIDIA}(2025)]{nvidia2025physicalai}
{NVIDIA}.
\newblock Physical {AI} autonomous vehicles dataset.
\newblock
  \url{https://huggingface.co/datasets/nvidia/PhysicalAI-Autonomous-Vehicles},
  2025.

\bibitem[{NVIDIA}(2026)]{nvidia2026alpamayo1.5}
{NVIDIA}.
\newblock Alpamayo-1.5.
\newblock \url{https://huggingface.co/nvidia/Alpamayo-1.5-10B}, 2026.

\bibitem[Paul et~al.(2024)Paul, West, Bosselut, and Faltings]{paul2024making}
Debjit Paul, Robert West, Antoine Bosselut, and Boi Faltings.
\newblock Making reasoning matter: Measuring and improving faithfulness of
  chain-of-thought reasoning.
\newblock In \emph{Findings of the Association for Computational Linguistics:
  EMNLP 2024}, pages 15012--15032, 2024.

\bibitem[{Qwen Team}(2026)]{qwen35}
{Qwen Team}.
\newblock {Qwen3.5-9B}.
\newblock \url{https://huggingface.co/Qwen/Qwen3.5-9B}, 2026.
\newblock Model card.

\bibitem[Ravi et~al.(2025)Ravi, Gabeur, Hu, Hu, Ryali, Ma, Khedr, R{\"a}dle,
  Rolland, Gustafson, Mintun, Pan, Alwala, Carion, Wu, Girshick, Doll{\'a}r,
  and Feichtenhofer]{ravi2025sam2}
Nikhila Ravi, Valentin Gabeur, Yuan-Ting Hu, Ronghang Hu, Chaitanya Ryali,
  Tengyu Ma, Haitham Khedr, Roman R{\"a}dle, Chloe Rolland, Laura Gustafson,
  Eric Mintun, Junting Pan, Kalyan~Vasudev Alwala, Nicolas Carion, Chao-Yuan
  Wu, Ross Girshick, Piotr Doll{\'a}r, and Christoph Feichtenhofer.
\newblock {SAM} 2: Segment anything in images and videos.
\newblock In \emph{International Conference on Learning Representations}, 2025.

\bibitem[Sima et~al.(2024)Sima, Renz, Chitta, Chen, Zhang, Xie, Bei{\ss}wenger,
  Luo, Geiger, and Li]{sima2024drivelm}
Chonghao Sima, Katrin Renz, Kashyap Chitta, Li~Chen, Hanxue Zhang, Chengen Xie,
  Jens Bei{\ss}wenger, Ping Luo, Andreas Geiger, and Hongyang Li.
\newblock {DriveLM}: Driving with graph visual question answering.
\newblock In \emph{European Conference on Computer Vision}, pages 256--274.
  Springer, 2024.

\bibitem[Song et~al.(2025)Song, Huai, Jiang, Kong, and Luo]{song2025more}
Xurui Song, Shuo Huai, JingJing Jiang, Jiayi Kong, and Jun Luo.
\newblock More than meets the eye? uncovering the reasoning-planning disconnect
  in training vision-language driving models.
\newblock \emph{arXiv preprint arXiv:2510.04532}, 2025.

\bibitem[Sun et~al.(2025)Sun, Cui, Dong, and Tang]{sun2025attentiveeraser}
Wenhao Sun, Benlei Cui, Xue-Mei Dong, and Jingqun Tang.
\newblock Attentive eraser: Unleashing diffusion model's object removal
  potential via self-attention redirection guidance.
\newblock In \emph{Proceedings of the AAAI Conference on Artificial
  Intelligence}, 2025.

\bibitem[Sun et~al.(2026)Sun, Zuo, Xie, Gao, Gong, and Ma]{sun2026facte}
Yuxi Sun, Aoqi Zuo, Haotian Xie, Wei Gao, Mingming Gong, and Jing Ma.
\newblock {FACT-E}: Causality-inspired evaluation for trustworthy
  chain-of-thought reasoning.
\newblock In \emph{Findings of the Association for Computational Linguistics:
  ACL 2026}, pages 20275--20293, 2026.

\bibitem[Tian et~al.(2024)Tian, Gu, Li, Liu, Wang, Zhao, Zhan, Jia, Lang, and
  Zhao]{tian2024drivevlm}
Xiaoyu Tian, Junru Gu, Bailin Li, Yicheng Liu, Yang Wang, Zhiyong Zhao, Kun
  Zhan, Peng Jia, Xianpeng Lang, and Hang Zhao.
\newblock {DriveVLM}: The convergence of autonomous driving and large
  vision-language models.
\newblock In \emph{Conference on Robot Learning}, pages 4698--4726. PMLR, 2024.

\bibitem[Turpin et~al.(2023)Turpin, Michael, Perez, and
  Bowman]{turpin2023language}
Miles Turpin, Julian Michael, Ethan Perez, and Samuel~R. Bowman.
\newblock Language models don't always say what they think: Unfaithful
  explanations in chain-of-thought prompting.
\newblock In \emph{Advances in Neural Information Processing Systems},
  volume~36, 2023.

\bibitem[Wang et~al.(2025{\natexlab{a}})Wang, Yu, Jiang, Lan, Shi, Chang,
  Kautz, Li, and Alvarez]{wang2025omnidrive}
Shihao Wang, Zhiding Yu, Xiaohui Jiang, Shiyi Lan, Min Shi, Nadine Chang, Jan
  Kautz, Ying Li, and Jose~M. Alvarez.
\newblock {OmniDrive}: A holistic vision-language dataset for autonomous
  driving with counterfactual reasoning.
\newblock In \emph{Proceedings of the IEEE/CVF Conference on Computer Vision
  and Pattern Recognition}, pages 22442--22452, 2025{\natexlab{a}}.

\bibitem[Wang et~al.(2025{\natexlab{b}})Wang, Luo, Bai, Cao, Che, Chen, Chen,
  Diamond, Ding, Ding, et~al.]{wang2025alpamayo}
Yan Wang, Wenjie Luo, Junjie Bai, Yulong Cao, Tong Che, Ke~Chen, Yuxiao Chen,
  Jenna Diamond, Yifan Ding, Wenhao Ding, et~al.
\newblock {Alpamayo-R1}: Bridging reasoning and action prediction for
  generalizable autonomous driving in the long tail.
\newblock \emph{arXiv preprint arXiv:2511.00088}, 2025{\natexlab{b}}.

\bibitem[Xu et~al.(2024)Xu, Zhang, Xie, Zhao, Guo, Wong, Li, and
  Zhao]{xu2023drivegpt4}
Zhenhua Xu, Yujia Zhang, Enze Xie, Zhen Zhao, Yong Guo, Kwan-Yee~K. Wong,
  Zhenguo Li, and Hengshuang Zhao.
\newblock {DriveGPT4}: Interpretable end-to-end autonomous driving via large
  language model.
\newblock \emph{IEEE Robotics and Automation Letters}, 9\penalty0
  (10):\penalty0 8186--8193, 2024.

\bibitem[Young(2026)]{young2026measuring}
Richard~J. Young.
\newblock Measuring faithfulness depends on how you measure: Classifier
  sensitivity in {LLM} chain-of-thought evaluation.
\newblock \emph{arXiv preprint arXiv:2603.20172}, 2026.

\bibitem[Yuan et~al.(2025)Yuan, Qian, Tang, Chen, Song, Sun, Chu, Cai, Zhang,
  and Li]{yuan2025autodrive}
Zhenlong Yuan, Chengxuan Qian, Jing Tang, Rui Chen, Zijian Song, Lei Sun,
  Xiangxiang Chu, Yujun Cai, Dapeng Zhang, and Shuo Li.
\newblock {AutoDrive-R$^2$}: Incentivizing reasoning and self-reflection
  capacity for {VLA} model in autonomous driving.
\newblock \emph{arXiv preprint arXiv:2509.01944}, 2025.

\bibitem[Zaman and Srivastava(2026)]{zaman2026really}
Kerem Zaman and Shashank Srivastava.
\newblock Is chain-of-thought really not explainability? chain-of-thought can
  be faithful without hint verbalization.
\newblock In \emph{Proceedings of the 64th Annual Meeting of the Association
  for Computational Linguistics (Volume 1: Long Papers)}, pages 48008--48030,
  2026.

\bibitem[Zhao et~al.(2026)Zhao, Wang, Yang, and Zhou]{zhao2026objectclear}
Jixin Zhao, Zhouxia Wang, Peiqing Yang, and Shangchen Zhou.
\newblock Precise object and effect removal with adaptive target-aware
  attention.
\newblock In \emph{Proceedings of the IEEE/CVF Conference on Computer Vision
  and Pattern Recognition}, 2026.

\bibitem[Zhou et~al.(2023)Zhou, Li, Chan, and Loy]{zhou2023propainter}
Shangchen Zhou, Chongyi Li, Kelvin~C.K. Chan, and Chen~Change Loy.
\newblock {ProPainter}: Improving propagation and transformer for video
  inpainting.
\newblock In \emph{Proceedings of the IEEE/CVF International Conference on
  Computer Vision}, 2023.

\bibitem[Zhou et~al.(2024)Zhou, Liu, Yurtsever, Zagar, Zimmer, Cao, and
  Knoll]{zhou2024vlmad}
Xingcheng Zhou, Mingyu Liu, Ekim Yurtsever, Bare~Luka Zagar, Walter Zimmer,
  Hu~Cao, and Alois~C. Knoll.
\newblock Vision language models in autonomous driving: A survey and outlook.
\newblock \emph{IEEE Transactions on Intelligent Vehicles}, pages 1--20, 2024.

\bibitem[Zhou et~al.(2025)Zhou, Cai, Zhao, Zhang, Huang, Zhou, and
  Ma]{zhou2025autovla}
Zewei Zhou, Tianhui Cai, Seth~Z. Zhao, Yun Zhang, Zhiyu Huang, Bolei Zhou, and
  Jiaqi Ma.
\newblock {AutoVLA}: A vision-language-action model for end-to-end autonomous
  driving with adaptive reasoning and reinforcement fine-tuning.
\newblock In \emph{Advances in Neural Information Processing Systems},
  volume~38, 2025.

\end{thebibliography}
}\end{document}